\pdfoutput=1

\documentclass[11pt]{article}
\usepackage{soul}
 \usepackage[final]{acl}

\usepackage{times}
\usepackage{latexsym}
\usepackage{booktabs}
\usepackage{multirow}
\usepackage{tabularx} 

\usepackage{longtable}

\usepackage{svg}
\usepackage{amsmath}

\usepackage[T1]{fontenc}

\usepackage[utf8]{inputenc}

\usepackage{microtype}

\usepackage{multirow}
\usepackage{xcolor}

\usepackage{inconsolata}
\usepackage{duckuments}

\usepackage{subfig}

\usepackage{comment}

\usepackage{graphicx}
\usepackage{xcolor}

\usepackage{cleveref}
\usepackage{enumitem}

\newcommand{\lee}[1]{{\color{red} [\bf {dongwon:}  #1]}}

\newcommand\thai[1]{\noindent{\color{purple} {\bf \fbox{Thai}} {\it#1}}}
\newcommand{\jc}[1]{{\color{blue} [\bf {JC:}  #1]}}

\title{{\sf PlagBench}: Exploring the Duality of Large Language Models in Plagiarism Generation and Detection}

\author{Jooyoung Lee$^1$\hspace{0.2in}
        Toshini Agrawal$^2$\hspace{0.2in}
        Adaku Uchendu$^3$ \\
        \textbf{Thai Le$^4$} \hspace{0.2in}
        \textbf{Jinghui Chen$^1$} \hspace{0.2in}
        \textbf{Dongwon Lee$^1$} \vspace{0.1in} \\
        $^1$ The Pennsylvania State University, University Park, PA, USA, 
        \texttt{\{jfl5838, jzc5917, dongwon\}@psu.edu} \\
        $^2$ Vellore Institute of Technology, Bhopal, India, 
        \texttt{agrawaltoshini@gmail.com} \\
        $^3$ MIT Lincoln Laboratory, Lexington, MA, USA, 
         \texttt{adaku.uchendu@ll.mit.edu} \\
        $^4$ Indiana University, Bloomington, IN, USA, 
        \texttt{tle@iu.edu}
        \vspace{0.1in} \\
        }

\begin{document}
\maketitle
\begin{abstract}
Recent studies have raised concerns about the potential threats large language models (LLMs) pose to academic integrity and copyright protection. Yet, their investigation is predominantly focused on literal copies of original texts. Also, how LLMs can facilitate the detection of LLM-generated plagiarism remains largely unexplored. To address these gaps, we introduce \textbf{{\sf PlagBench}}, a dataset of 46.5K synthetic text pairs that represent three major types of plagiarism: verbatim copying, paraphrasing, and summarization. These samples are generated by three advanced LLMs. We rigorously validate the quality of \textbf{{\sf PlagBench}} through a combination of fine-grained automatic evaluation and human annotation. We then utilize this dataset for two purposes: (1) to examine LLMs' ability to transform original content into accurate paraphrases and summaries, and (2) to evaluate the plagiarism detection performance of five modern LLMs alongside three specialized plagiarism checkers. Our results show that GPT-3.5 Turbo can produce high-quality paraphrases and summaries without significantly increasing text complexity compared to GPT-4 Turbo. However, in terms of detection, GPT-4 outperforms other LLMs and commercial detection tools by 20\%, highlights the evolving capabilities of LLMs not only in content generation but also in plagiarism detection. Data and source code are available at \url{https://github.com/Brit7777/plagbench}.


\end{abstract}

\section{Introduction}
Plagiarism occurs when someone uses another person's work, ideas, or expressions without proper acknowledgment. It is one of the most frequently scrutinized issues in writing tasks, which leads to a violation of intellectual property rights and academic integrity. Traditionally, plagiarism research has focused on human-written text. However, the rise of large language models (LLMs) has transformed the landscape of plagiarism \cite{pudasaini2024survey}. For instance, users can exploit these models to obfuscate original content through paraphrases or summaries, effectively bypassing traditional plagiarism detection systems \cite{krishna2024paraphrasing, sadasivan2023can, weber2023testing, elkhatat2021some}. Moreover, many recent studies \cite{carlini2021extracting, tirumala2022memorization, zhang2023counterfactual, carlini2023quantifying} have shown that LLMs can memorize portions of their training data and reproduce them during text generation. These models are not limited to producing verbatim copies; they can also generate paraphrased or elongated versions of unique content \cite{lee2023language}. 




\begin{figure}  
    \centering
    \includegraphics[width=1. \linewidth]{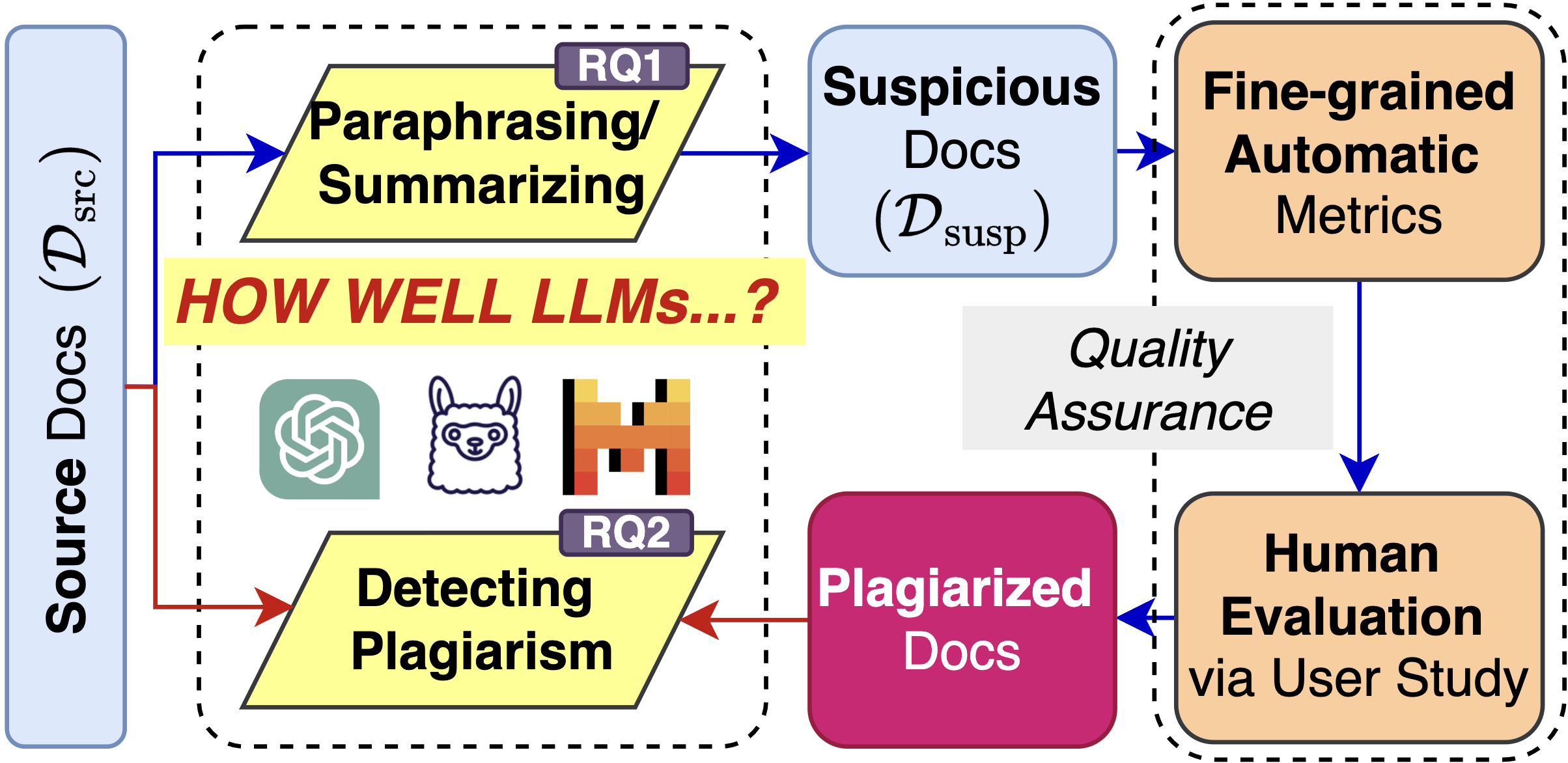}
    \caption{The overview of {\sf PlagBench} construction processes and proposed RQs. \textcolor{blue}{Blue} and \textcolor{red}{red} arrows denote the flow of \textit{RQ1} and \textit{RQ2}, respectively.}
    \label{fig:teaser} 
\end{figure}


The increasing prevalence of plagiarism in LLMs raises two critical research questions: \textit{(RQ1) how well can LLMs generate paraphrase and summary plagiarism of given texts?} and \textit{(RQ2) how well can LLMs detect three types of plagiarism--i.e., verbatim, paraphrase, and summary plagiarisms?}. While prior research has explored LLMs' applications to automated paraphrase \cite{witteveen-andrews-2019-paraphrasing, wahle-etal-2022-large} and summary \cite{lingard2023writing, goyal2022news, liu-etal-2023-revisiting} generation, these studies are often limited in scope because most of them focus on individual domains or specific models. A holistic and fine-grained analysis of diverse domains and models is crucial to developing robust mechanisms for detecting and mitigating plagiarism in AI-generated content. Additionally, LLMs have demonstrated impressive performance in various natural language processing (NLP) tasks \cite{chang2024survey}, motivating us to investigate their potential in automatic plagiarism detection. Current plagiarism detectors are primarily optimized to catch literal copies or superficial paraphrasing by humans and may not be sufficient to detect nuanced rewritings of LLMs. Evaluating LLMs' effectiveness in plagiarism detection can offer critical insights into the limitations of current systems and help guide the development of more advanced, robust solutions.

To answer these questions, we present \textbf{\sf PlagBench}, a large-scale corpus containing 46.5K text pairs with three plagiarism categories: verbatim, paraphrase, and summary plagiarism. These samples cover three writing domains (i.e., abstract of scholarly paper, story, news article) and are synthetically generated by three advanced, closed-source and open-source LLMs (Llama2-70b-chat \cite{touvron2023llama}, GPT-3.5 Turbo\footnote{\url{https://platform.openai.com/docs/models/gpt-3-5-turbo}}, and GPT-4 Turbo \cite{achiam2023gpt}). To acquire the gold standard for automatic plagiarism detection tasks, we employ a comprehensive integration of both automatic and manual evaluation. This final subset is then used for evaluation of five popular LLMs (Llama2-70b-chat, Llama3-70b-instruct\footnote{\url{https://ai.meta.com/blog/meta-llama-3/}}, Mixtral-8x7B \cite{jiang2024mixtral}, GPT-3.5 Turbo, and GPT-4 Turbo) and three specialized plagiarism detectors (GPTZero\footnote{\url{https://gptzero.me/plagiarism-checker}}, Prepostseo\footnote{\url{https://www.prepostseo.com/plagiarism-checker}}, and a detector proposed by \citet{lee2023language}), two of which are commercial-ready. Figure \ref{fig:teaser} illustrates a visual representation of our dataset construction processes and the proposed RQs.


Key contributions of this work are as follows: (1) \textbf{\sf PlagBench} is the first dataset that covers a wide range of plagiarism examples generated by state-of-the-art LLMs; (2) we showcase the strengths and weaknesses of LLM-generated plagiarism cases across domains; (3) our results show the potential of LLMs facilitating plagiarism detection, although they generally have difficulties in distinguishing the type of summary plagiarism.

\section{Related Work}

\subsection{Plagiarism Detection Corpus}
The Microsoft Research Paraphrase Corpus (MRPC) \cite{dolan2005automatically} and PAWS \cite{paws2019naacl} are among the most renowned datasets for paraphrase identification. Yet, they are not suitable for plagiarism detection tasks because they only cover one type. Additionally, their samples are annotated at the sentence level, which limits their applicability in real-world scenarios where plagiarism in general occurs at the paragraph level. More recently, \citet{Wahle2021} constructed a paragraph-long machine-paraphrased plagiarism dataset using three neural language models such as BERT \cite{Devlin2019BERTPO}, RoBERTa \cite{liu2019roberta}, and Longformer \cite{Beltagy2020Longformer}. Following, the authors replaced these three models with more recently released LLMs, GPT-3, and T5, and produced machine-paraphrased texts \cite{wahle-etal-2022-large}. There are many independent machine-summarized text pairs (e.g., \citet{cohan-etal-2018-discourse, xsum-emnlp}) as well, but there exists no comprehensive dataset covering both paraphrases and summaries generated by modern LLMs.

\citet{potthast2013overview} introduced a novel plagiarism detection dataset with five obfuscation strategies: no obfuscation, random obfuscation, translation obfuscation, and summary obfuscation. These strategies utilized tools like Google Translate and text summarizers to mimic plagiarist behaviors. To the best of our knowledge, this is the first and only plagiarism detection dataset covering a wide range of plagiarism types. However, this dataset has not been updated since 2013.







\subsection{Automatic Plagiarism Detection}

There are two formats of the automated plagiarism detection task: \textit{intrinsic} and \textit{extrinsic} plagiarism detection. While intrinsic plagiarism detection analyzes the document itself without resorting to the reference document, extrinsic plagiarism detection involves directly comparing the suspicious document against the reference to detect plagiarism instances \cite{alzahrani2011understanding}. Intrinsic approaches use patterns such as linguistic differences within a document which indicates multiple authorship to detect plagiarism \cite{gipp2014citation,xiong2024efficient,potthast2010evaluation, song2024programming,stein2011intrinsic}. Due to the high complexity of intrinsic plagiarism tasks, many popular detection tools follow extrinsic plagiarism detection strategies. The extrinsic plagiarism detection task is generally divided into two subtasks: source retrieval (i.e., finding the most plausible pair of the source and the suspicious document) and plagiarism identification (i.e., classifying whether two documents are plagiarizing each other or not). These approaches rely on similarity scores of two non-zero vectors transformed from the document pairs and apply certain thresholds
\cite{vani2017detection,sauglam2024automated,potthast2010evaluation,sauglam2024detecting,lee2023review,moravvej2023novel,avetisyan2023cross}. 
Our focus lies in the latter scope. We assume we already have the text pair, the source document, and the suspicious document obtained through \textbf{\sf PlagBench}, and we concentrate on plagiarism identification.

\section{\textbf{\sf PlagBench} Corpus Generation}


\subsection{Corpus Construction}

\noindent \textbf{Source Document Collection.}\label{subsec_doc_collection}
Although plagiarism can occur in any type of writing, we target specific areas such as scholarly or creative works, where originality and intellectual property are highly valued. \citet{li2023deepfake} present a collection of human-authored and machine-authored corpora from diverse writing tasks. Among their selections of corpora, we carefully choose three public English datasets: (1) \textbf{SciXGen} \cite{chen2021scixgen} consisting of 200K+ abstracts of scientific articles (\textit{for scientific writing}), (2) \textbf{ROCStories} \cite{mostafazadeh2016corpus} consisting of 50k five-sentence commonsense stories (\textit{for story writing}); and (3) \textbf{TLDR}\footnote{\url{https://huggingface.co/datasets/JulesBelveze/tldr_news}} consisting of 7K+ TLDR tech newsletters (\textit{for news article writing}). Their samples contain one human-written text and multiple machine-generated variations, all sharing either the same topic or text source as the human-written piece. For the purpose of the study, we only use the human-written text containing at least 5 sentences as the source text.\footnote{According to our pilot experiment, paraphrases and summaries generated based on documents shorter than 5 sentences tend to be too similar to distinguish even from the human lens.} We randomly select 2,400 samples from SciXGen and ROCStories corpora and 1,300 from TLDR for suspicious document generation. The sample size of TLDR is smaller than the other two datasets because the content of TLDR tends to be short.

\noindent \textbf{Suspicious Document Generation.} \label{subsec_susp_doc_gen}
Instruction-tuned LLMs have demonstrated strong abilities in text generation that are aligned with users' natural language commands \cite{lingard2023writing, zhang2024benchmarking, tang2023evaluating}. Hence, we employ Llama2-70b-chat, GPT-3.5 Turbo, and GPT-4 Turbo models to generate texts corresponding to two plagiarism labels (i.e., paraphrase and summary plagiarism), respectively. We do not use these models for verbatim plagiarism, as generating suspicious documents can be easily accomplished by simple copy-and-paste without any modification. Table \ref{tab:prompt_template_gen} shows our hand-crafted prompt templates, each tailored to transform a single source document according to the provided task descriptions. 

In plagiarism-free scenarios, existing datasets (e.g., \citet{potthast2013overview}, \citet{foltynek2020detecting}) pair random or dissimilar documents. However, this often results in pairs that are clearly different and thus easily identifiable. More challenging pairs are likely to occur when two documents share the same key topics and domain type, but their details do not overlap. Consequently, we instruct LLMs to write continuations based on provided keywords, domain information, and the first two sentences of the source document. Two meaningful keywords are automatically extracted using KeyBERT,\footnote{\url{https://maartengr.github.io/KeyBERT/index.html}} a package leveraging contextual embeddings from BERT \cite{devlin-etal-2019-bert}. 

\noindent \textbf{Generation Results.} We have in total 6,100 source documents across three writing domains. Using three LLMs, we create three corresponding paraphrase, summary, and plagiarism-free documents per sample. For GPT-4, due to its relatively high API cost\footnote{Llama2 families are open-source models while GPT-3.5 turbo and GPT-4 Turbo charge \$3.00 and \$40.00 per 1M tokens respectively.}, we only utilize a subset (n=3,300) of source documents for generation. We use the version of \textit{gpt-3.5-turbo-1106} and \textit{gpt-4-1106-preview} and apply temperature sampling (\textit{t} = 0.9) for the generation configuration. As a result, the total number of generated suspicious documents is 46,500, a summation of 6,100 generations for SciXGen, 6,100 generations for ROCStories, and 3,300 generations for TLDR, multiplied by 3 plagiarism types.

\begin{table*}[]
\footnotesize{
\setlength{\tabcolsep}{0.5em} 
{\renewcommand{\arraystretch}{1.3}
\begin{tabular}{|c|c|l|}
\hline
\textbf{Type} &
  \textbf{Metric} &
  \multicolumn{1}{c|}{\textbf{Description}} \\ \hline
\multirow{3}{*}{\textbf{Paraphrase}} &
  \begin{tabular}[c]{@{}c@{}}Semantic \\ equivalence\end{tabular} &
  Evaluates whether the paraphrased text conveys the same meaning as the source text. \\ \cline{2-3} 
 &
  Consistency &
  \begin{tabular}[c]{@{}l@{}}Evaluates whether the paraphrased text accurately represents the information contained\\ in the source text without the inclusion of errors or distortions.\end{tabular} \\ \cline{2-3} 
 &
  \begin{tabular}[c]{@{}c@{}}Language\\  quality\end{tabular} &
  \begin{tabular}[c]{@{}l@{}}Evaluates whether the paraphrased text maintains or improves upon the quality of the\\ source text in regards to fluency and grammaticality.\end{tabular} \\ \hline
\multirow{4}{*}{\textbf{Summary}} &
  Relevance &
  \begin{tabular}[c]{@{}l@{}}Evaluates whether the summary covers all the essential information and key points from \\ the source text while omitting irrelevant or tangential details.\end{tabular} \\ \cline{2-3} 
 &
  Coherence &
  \begin{tabular}[c]{@{}l@{}}Evaluate whether the summary is well-structured and not just a random assortment of \\ information from the source text.\end{tabular} \\ \cline{2-3} 
 &
  Consistency &
  \begin{tabular}[c]{@{}l@{}}Evaluates whether the summary accurately represents the content and meaning of the\\ source text without the inclusion of errors or distortions.\end{tabular} \\ \cline{2-3} 
 &
  \begin{tabular}[c]{@{}c@{}}Language \\ quality\end{tabular} &
  \begin{tabular}[c]{@{}l@{}}Evaluates whether the summary maintains or improves upon the quality of the source \\ text in regards to fluency and grammaticality.\end{tabular} \\ \hline
\end{tabular}
}
}
\caption{Descriptions of evaluation metrics for paraphrase and summary plagiarism.}
\label{tab:evaluation_metrics}
\end{table*}

\subsection{Corpus Quality Assurance via Automated Metrics}

\noindent \textbf{Automatic Evaluation.}
\label{sec:automatic_eval}
Hallucination in LLMs is a widely recognized problem. To ensure that their generations are accurate paraphrases and summaries, we rigorously evaluate them based on our established metrics (Table  \ref{tab:evaluation_metrics}) and apply filtering strategies. Due to relatively large volumes of generated samples, we first conduct automatic measurements for paraphrase and summary plagiarism. Note that we do not perform automatic evaluation for non-plagiarism cases because it would overlap with our benchmark task and necessitate the use of the same plagiarism detectors we aim to evaluate. 

For paraphrase plagiarism, we perform an evaluation of LLM-paraphrased documents through the lens of semantic equivalence, consistency, and language quality. Several papers \cite{shen2022evaluation, chen2011collecting} adopted similar criteria for paraphrase evaluation. We use the source documents as reference documents and carefully choose a single representative metric for every aspect.  
\begin{itemize}[leftmargin=\dimexpr\parindent-0.1\labelwidth\relax,noitemsep]
    \item \textbf{Semantic equivalence} $\rightarrow$ BERTScore \cite{zhang2019bertscore} : BERTScore retrieves the token-level embedding using BERT and computes the summation of cosine similarities. It is shown to be more robust for paraphrase classification than other conventional metrics like BLEU \cite{papineni2002bleu} and METEOR \cite{banerjee2005meteor}. 
    \item \textbf{Consistency} $\rightarrow$ AlignScore \cite{zha-etal-2023-alignscore} : the authors train a information alignment model on 7 well-established tasks including paraphrasing and summarization.
    \item \textbf{Language quality} $\rightarrow$ Readability index  \cite{senter1967automated} \& CoLAScore \cite{zhu-bhat-2020-gruen} : we use automated readability index for fluency and use COLAScore, a RoBERTa-based model finetuned on grammaticality measurement data, for grammaticality measurement.
\end{itemize}

Now we turn our attention to automatic summary evaluation. Specifically, we measure relevance, coherence, consistency, and language quality, suggested by the SummEval benchmark \cite{fabbri2021summeval}. In the absence of gold-standard summaries for comparison with LLM-generated summaries, we rely on reference-free metrics. Below are the descriptions of selected metrics:
\begin{itemize}[leftmargin=\dimexpr\parindent-0.1\labelwidth\relax,noitemsep]
\item \textbf{Relevance} $\rightarrow$ BLANC \cite{vasilyev-etal-2020-fill} : BLANC is a novel approach that calculates the usefulness of a summary in helping BERT for language understanding task. This metric is ranked the highest in relevance alignment in SummEval. 
\item \textbf{Coherence} $\rightarrow$ BARTScore \cite{NEURIPS2021_e4d2b6e6} : BARTScore is a comprehensive evaluation metric using a pre-trained sequence-to-sequence (seq2seq) model BART \cite{lewis-etal-2020-bart}.
\item We use the same models from paraphrase evaluation regarding consistency and language quality.
\end{itemize}

Most metrics, except for CoLAScore, provide continuous values, requiring threshold selection for filtering. The primary goal of automated evaluation is to eliminate inaccurate pairs. Thus, we shuffle the order of suspicious document rows while keeping the source document order intact and create invalid pairs as a means of establishing a loose cut-off point. Refer to Appendix \ref{sec:appendix_threshold} for experiment configurations and the threshold setup. Lastly, we remove near-duplicates based on the Levenshtein distance \cite{miller2009levenshtein} between two texts.

\noindent \textbf{Automatic Filtering Results.} We remove samples from a final set if the text pair does not satisfy all 3 or 4 aspects, depending on the plagiarism category. Among 31,000 (15,500 per plagiarism type) suspicious documents, 12,071 samples for paraphrase plagiarism and 13,445 samples for summary plagiarism remain intact. The document percentage breakdown per plagiarism type and domain type after automatic filtering is illustrated in Table \ref{tab:filter_stats}.

\subsection{Corpus Quality Assurance via Manual Annotation}
Human annotation plays a critical role in verifying the accuracy and reliability of automated systems. We use Amazon Mechanical Turk (AMT) to hire human annotators for three separate annotation tasks: non-plagiarism evaluation, paraphrase evaluation, and summary evaluation. See Appendix \ref{sec:appendix_amt} for more details on the experiment design.
Due to limited budgets, we perform manual annotation on 459 batches (i.e., 1,377 samples) for no plagiarism, 393 batches (i.e., 1,251 samples) for paraphrase plagiarism, and 417 batches (i.e., 1,251 samples) for summary plagiarism cases that are randomly sampled from \S \ref{sec:automatic_eval}. 

We take several actions to rigorously reinforce annotation qualities. First, we employ four built-in worker qualifications, including (1) HIT Approval Rate of ${\leq}98\%$, (2) a minimum of 1,000 Approved HITs, (3) Masters qualification status, and (4) U.S-based workers. Second, a honey pot question is included to detect automated responses or bots. Third, to ensure that annotators spend enough time to carefully read the provided texts, we systematically block the `submit' button until 10 minutes have been reached. Lastly, we purposely add three dummy questions consisting of irrelevant text pairs. These questions serve to filter the annotators who do not fully understand the task or submit random responses without reading. We repeat this procedure until all batches have three approved responses. 

Each sample from our final annotation consists of three labels rated by multiple annotators. To measure the consistency and reliability of their evaluation, we compute the inter-rater agreement scores by counting the proportion of labels that all annotators agreed to. The scores are as follows: 87.15\% for plagiarism-free cases, 51.64\% for paraphrase plagiarism cases, and 75.32\% for summary plagiarism cases. The relatively low score for paraphrase plagiarism may be explained by the complexity and subjectivity involved in identifying paraphrased content, as human raters may disagree on how much rewording is acceptable to preserve semantic equivalence. Based on the manual inspection, when the LLM-generated text does not have many word overlaps with the source text with increased complexity, the agreement rate is prone to diverge. To address this misalignment, we filter out those samples with below 50\% agreement scores. 

\noindent \textbf{Manual Filtering Results.} After filtering samples with low inter-rater agreement scores, the final data contains 1,239, 545, and 1,181 samples for the plagiarism-free (94.95\% agreement rates), paraphrase (88.01\% agreement rates), and summary (85.78\% agreement rates) plagiarism labels. We take the majority votings of three annotators' responses and consider them as gold standard labels per evaluation metrics. Consistent with automatic evaluation filtering, we further remove samples that fail to satisfy all 3 or 4 aspects, depending on the plagiarism category. Table \ref{tab:human_filter_stats} illustrates the document percentage breakdown per plagiarism type and domain type after manual filtering.

\begin{figure}[]
  \centering
  \includegraphics[width=\linewidth]{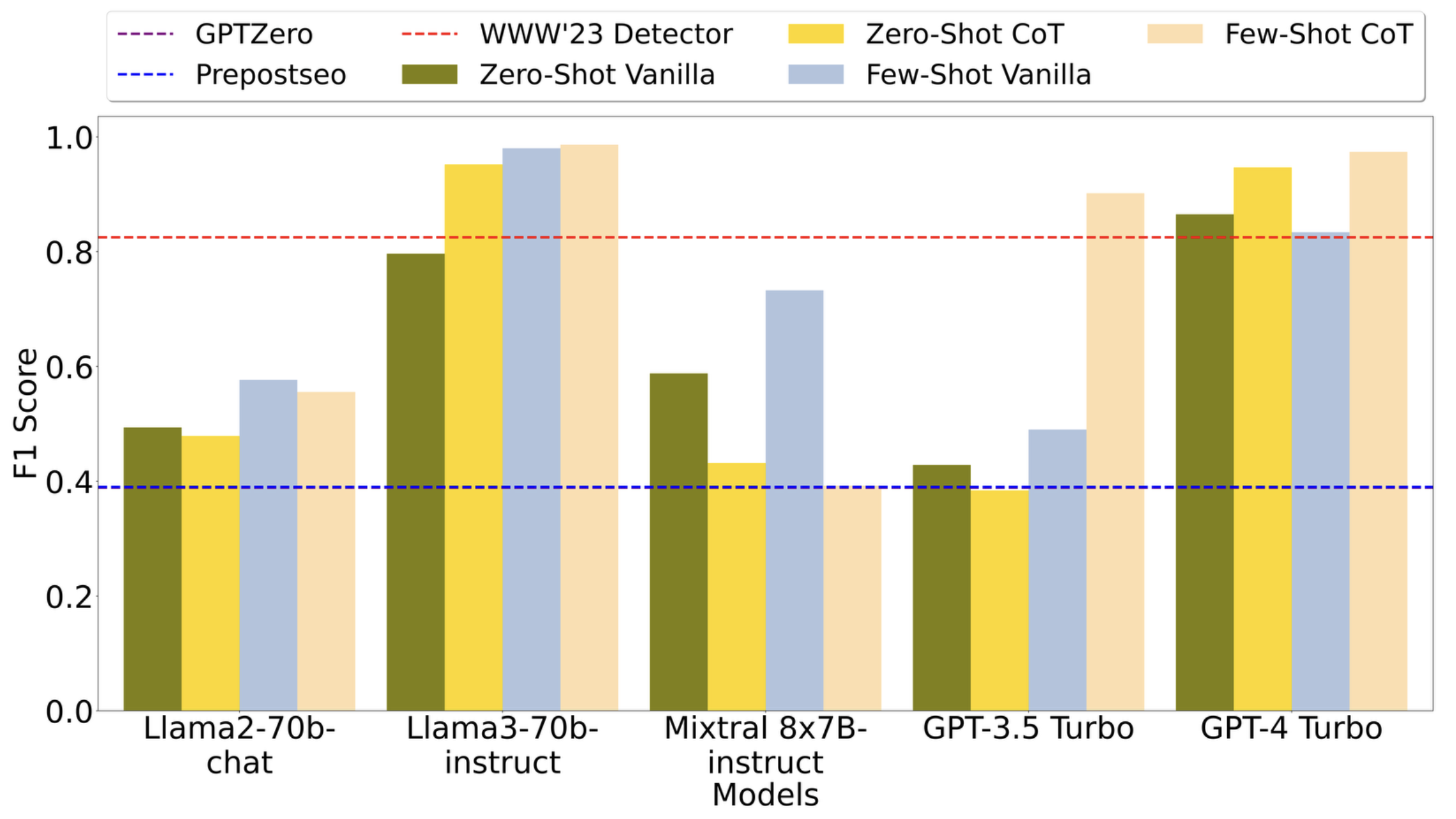}
    \caption{Binary plagiarism detection (no plagiarism vs. plagiarism) performance of 5 LLMs w.r.t. prompt types. Dotted lines represent the performance of non-LLM based detectors.} 
    \label{fig:binary_perf}
  \vspace{-0.5cm}
\end{figure}

\section{\textbf{\sf PlagBench} Plagiarism Detection Task}

\textbf{Taxonomy of Plagiarism.} we focus on detection of extrinsic plagiarism detection that involves directly comparing a suspicious document $D_{susp}$ against a source document $D_{src}$ to detect plagiarism instances \cite{alzahrani2011understanding}. Following the plagiarism detection task proposed by PAN,\footnote{\url{https://pan.webis.de}} we study three branches of plagiarism: \textit{verbatim}, \textit{paraphrase}, and \textit{summary}. 
Verbatim plagiarism occurs when exact copies of words or phrases from $D_{src}$ are found in $D_{susp}$ without quotation marks. Paraphrase plagiarism, on the other hand, is when $D_{susp}$ rephrases $D_{src}$ using different words but retaining the same meaning and structure without providing a citation. Synonymous substitution, word reordering, and insertion/deletion are common paraphrasing techniques used by plagiarists \cite{alvi2021paraphrase}. While both verbatim and paraphrase plagiarism cases tend to maintain the same length and structure as the original text, summary plagiarism involves succinctly summarizing $D_{src}$ to reuse its key points or ideas. As summaries leave out unnecessary information, they are prone to be significantly shorter than $D_{src}$. Unlike verbatim plagiarism, which can be effectively captured using simple string-matching algorithms, paraphrase, and summary plagiarism are more challenging to identify due to their limited lexical and syntactic similarities to the original text. Specifically, these two categories require careful attention to the meaning and context of the content rather than just high-level vocabulary overlaps. 

\noindent \textbf{Task Formulation.} To investigate LLMs' capabilities in plagiarism detection, we present two tasks: 1) binary classification (plagiarism-free vs. plagiarism) and 2) plagiarism type identification (plagiarism-free vs. verbatim vs. paraphrase vs. summary plagiarism). We hypothesize that the second task is more difficult than the first task because the detector must capture more nuanced differences among the types of plagiarism. Our test set consists of 810 pairs of $D_{susp}$ and $D_{susp}$ randomly sampled from human annotation results. Specifically, it has 45 text pairs from no plagiarism labels and 45 from three plagiarism labels, generated by three LLMs within three writing domains. 

\noindent \textbf{Experiment Setup.} Using the 810 test set, we evaluate Five LLMs, Llama2-70b-chat, Llama3-70b-instruct, Mixtral-8x7B-instruct, GPT-3.5 Turbo, and GPT-4 Turbo, for binary classification of plagiarism and plagiarism type identification. In addition, three non-LLM-based plagiarism detectors are included for evaluation. Two of these, GPTZero\footnote{\url{https://gptzero.me/}} and Prepostseo,\footnote{\url{https://www.prepostseo.com/plagiarism-checker}} are commercial plagiarism checkers, while the third is a detector proposed by \citet{lee2023language}. We denote \citet{lee2023language}'s detector as the WWW'23 detector for the rest of the paper. All these tools rely on semantic similarity measures and text alignment algorithms to distinguish plagiarism. 

For LLM evaluation, we experiment with four popular prompting techniques: zero-shot vanilla prompting, zero-shot Chain-of-Thought (CoT) prompting \cite{wei2022chain}, few-shot vanilla prompting, and few-shot CoT prompting (Table \ref{tab:prompt_example}). CoT prompting is one of the most popular prompt techniques to enhance models' downstream task performance by eliciting reasoning. As shown in Table \ref{tab:prompt_example}, we define the definitions of plagiarism categories inside the prompt, aiming to enhance the alignment of LLMs with our task. For few-shot experiments, we provide six demonstrations, a mix of three plagiarism pairs and three plagiarism-free pairs. We leverage GPT-4 to obtain high-quality reasoning for few-shot CoT prompting. For generation, we employ greedy decoding (i.e., choosing the most plausible token at each generation step) to support the reproducibility of our results to some extent. Finally, we report F1 scores for detection performance.

\begin{figure}[]
  \centering
  \includegraphics[width=\linewidth]{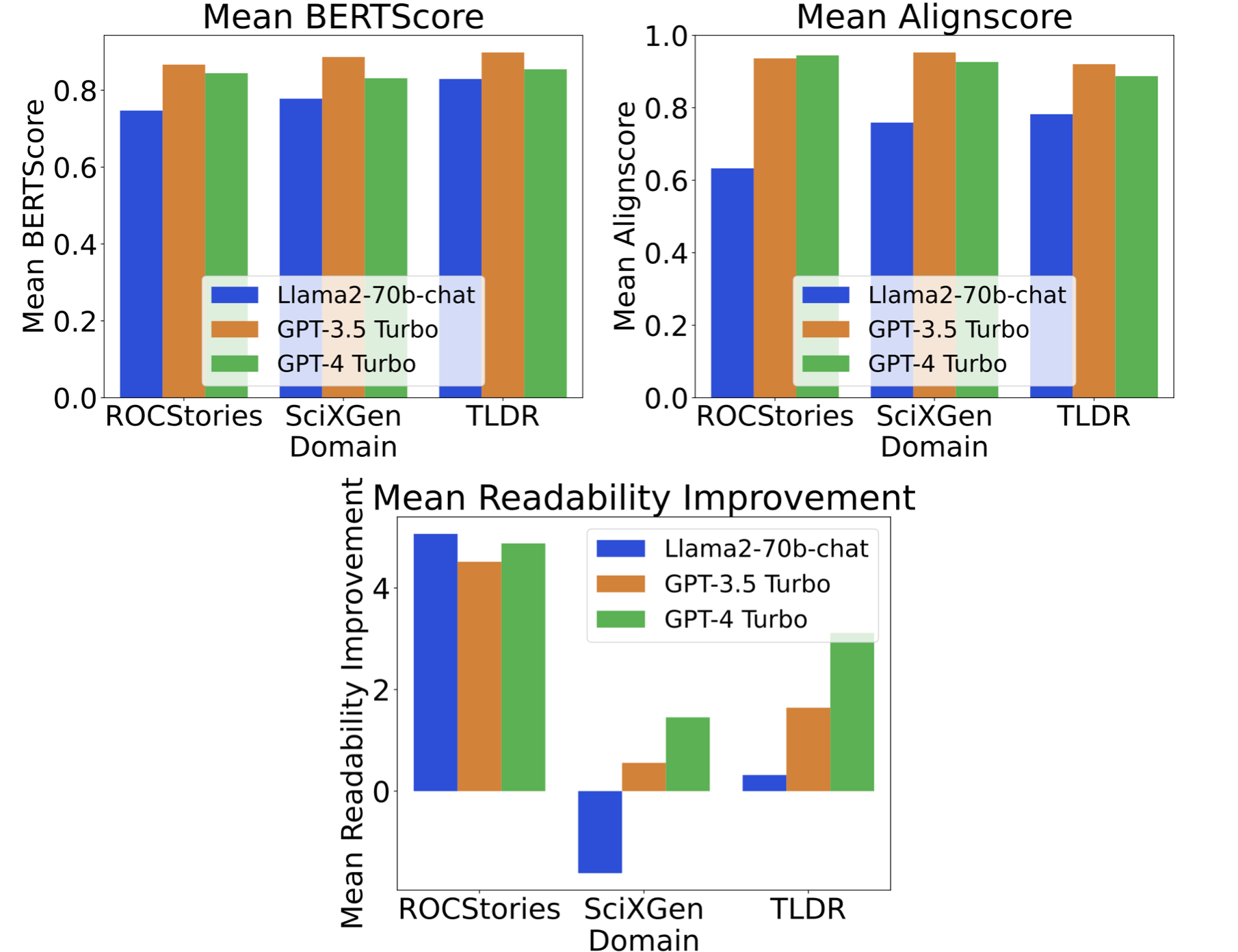}
    \caption{Mean paraphrase evaluation aspect scores w.r.t. domain and model types (automatic evaluation). For BERTScore and Alignscore, a higher score indicates greater semantic similarity and consistency between the LLM-paraphrased text and the source text. A higher readability improvement score suggests that the LLM-paraphrased text is more complex and requires a higher level of education to understand compared to the source text.} 
    \label{fig:automatic_para_eval}
  \vspace{-0.5cm}
\end{figure}

\begin{figure}[]
  \centering
  \includegraphics[width=\linewidth]{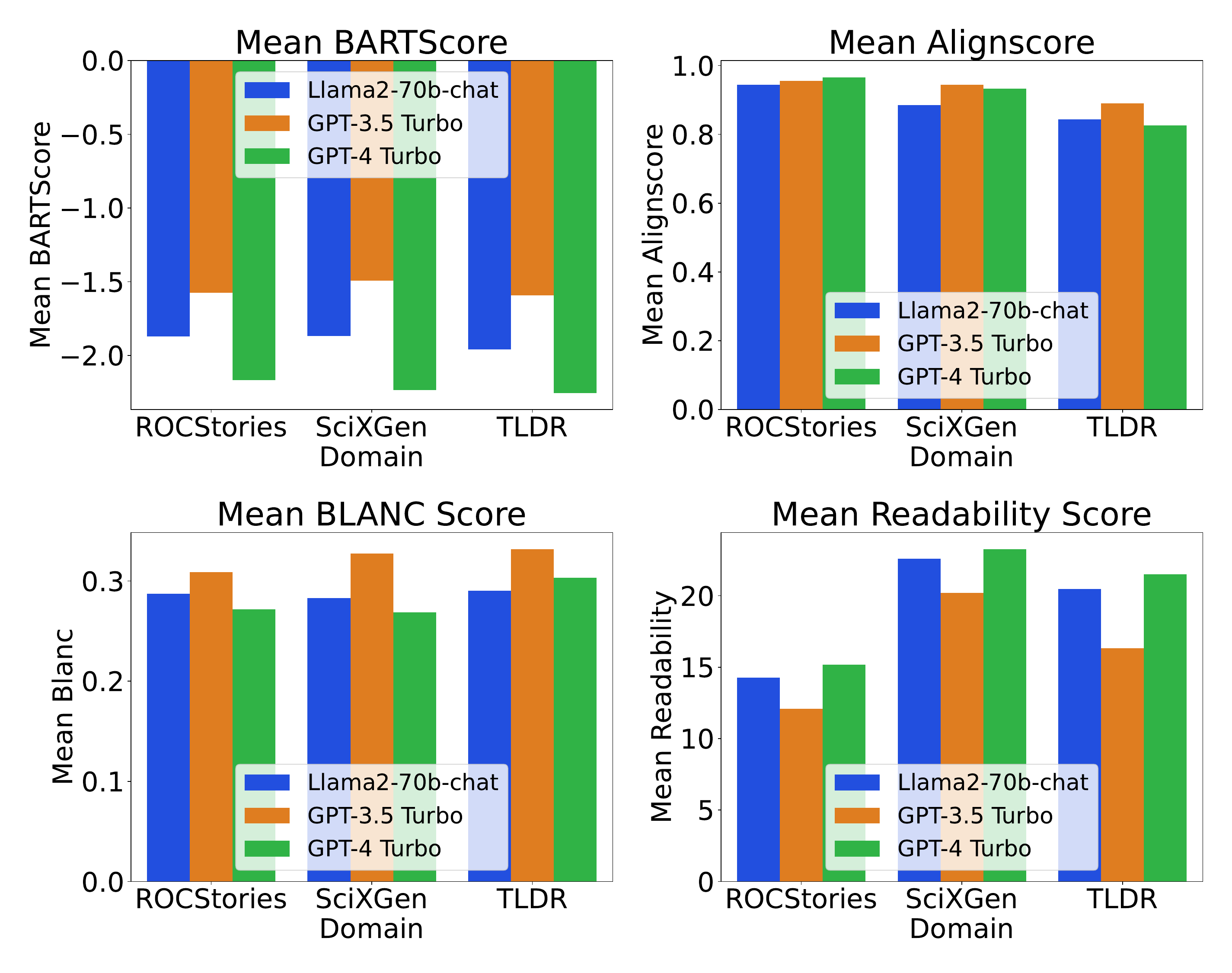}
    \caption{Mean summary evaluation aspect scores w.r.t. domain and model types (automatic evaluation). The lower the BARTScore the more coherent given a pair of the source text and the LLM-summarized text. For Alignscore and BLANC score, a higher score indicates greater consistency and relevancy between the LLM-summarized text and the source text. A higher readability score suggests that the LLM-summarized text is more complex and requires a higher level of education to understand compared to the source text.} 
    \label{fig:automatic_summary_eval}
  \vspace{-0.5cm}
\end{figure}

\section{Experiment Results}
\subsection{RQ1. how well can LLMs generate paraphrase and summary plagiarism?} 

\vspace{5pt}
\noindent \textbf{Automatic Paraphrase Evaluation.} The purpose of this analysis is to comprehend how accurately Llama2-70b-chat, GPT-3.5 Turbo, and GPT-4 Turbo can perform paraphrasing and summarization. Hence, we examine automated evaluation scores of their generations \textit{prior to} data filtering. 
Figure \ref{fig:automatic_para_eval} shows mean evaluation scores with respect to types of models and writing domains in paraphrase generation. We exclude the grammatical analyses since the majority of LLM-reworded texts are grammatically correct. The results indicate that Llama2-70b-chat achieves significantly lower performance in BERTScore and Alignscore compared to both GPT-3.5 Turbo and GPT-4 Turbo. The observed pattern persists across three writing domains. Particularly, Llama2-70b-chat suffers from generating factually consistent paraphrases for short stories (ROCStories). This finding is consistent with existing literature \cite{mishra2024fine, dahl2024large} highlighting the vulnerability of Llama2 in hallucination. Within the scope of semantic equivalence and consistency metrics, GPT-3.5 Turbo achieves the highest score, even occasionally better than GPT-4 Turbo. Analyses on readability score\footnote{We use the automated readability index to measure readability. A higher readability index hints that the text is more complex and may necessitate a higher level of education or reading proficiency to comprehend.} gaps between the human-written source text and the machine-generated suspicious text suggest that all three models significantly complicate ROCStories during rephrasing. In regards to SciXGen and TLDR, GPT models tend to produce more sophisticated texts, whereas Llama2-70b-chat often degrades linguistic quality. Substantial readability gaps, especially in the GPT-4 version, are somewhat anticipated, as \citet{onder2024evaluation, momenaei2023appropriateness} have noted that ChatGPT's outputs can be challenging to understand, often demanding a college-level proficiency in linguistic skills.

\vspace{5pt}
\noindent \textbf{Automatic Summary Evaluation.}  Summary evaluation results are illustrated in Figure \ref{fig:automatic_summary_eval}. Regardless of writing domain categories, GPT-3.5 Turbo is ranked the lowest in BARTScore, followed by  Llama2-70b-chat and then GPT-4 Turbo. The lower the BARTScore the more coherent given a pair of the source document and the machine-summarized document. There are no noticeable differences between the three models regarding the Alignscore, suggesting that they excel at producing factually consistent summaries. Still, GPT-3.5 Turbo outperforms both models in the domains of SciXGen and TLDR. Also, BLANC scores are shown to be the highest for GPT-4 Turbo while GPT-4 Turbo was ranked the lowest. The higher the BLANC score is the more relevant and useful the given summary is. As opposed to paraphrase analyses which take into account the difference between two pieces of text, here we only compute the readability index scores of LLM-generated summaries. This is because we do not have reference summaries to compare against. Consistent with paraphrase generation, GPT-4 Turbo tends to produce summaries with higher readability scores in comparison to Llama2-70b-chat and GPT-3.5 Turbo. Interestingly, summaries generated by Llama2-70b-chat are associated with higher readability scores than GPT-3.5 Turbo. These findings overall suggest that GPT-3.5 Turbo is generally the best paraphraser and summarizer in terms of all aspects, except for fluency/readability. Also, while Llama2-70b-chat is ranked the worst model for paraphrasing tasks, GPT-4 Turbo suffers the most for providing high-quality summaries.

\begin{table*}[tb]
\resizebox{\textwidth}{!}{%
\centering
\begin{tabular}{@{}|c|c|ccc|ccc|ccc|c|@{}}
\toprule
 &
  N/A &
  \multicolumn{3}{c|}{Llama2-70b-chat} &
  \multicolumn{3}{c|}{GPT-3.5 Turbo} &
  \multicolumn{3}{c|}{GPT-4 Turbo} &
   \\ \cmidrule(lr){2-11}
\multirow{-2}{*}{\textbf{Detectors}} &
  Verb &
  No &
  Para &
  Summ &
  No &
  Para &
  Summ &
  No &
  Para &
  Summ &
  \multirow{-2}{*}{Acc} \\ \midrule
\begin{tabular}[c]{@{}c@{}}Llama2-70b-chat\end{tabular} &
  66.67\% &
  36.29\% &
  68.88\% &
  6.66\% &
  30.37\% &
  55.55\% &
  8.88\% &
  33.33\% &
  71.11\% &
  0\% &
  39.5\% \\
\begin{tabular}[c]{@{}c@{}}Llama3-70b-instruct\end{tabular} &
  {\color[HTML]{CB0000} \textbf{100\%}} &
  85.92\% &
  {\color[HTML]{CB0000} \textbf{95.55\%}} &
  11.11\% &
  81.48\% &
  {\color[HTML]{CB0000} \textbf{100\%}} &
  11.11\% &
  78.51\% &
  {\color[HTML]{CB0000} \textbf{100\%}} &
  8.88\% &
  75.06\% \\
\begin{tabular}[c]{@{}c@{}}Mixtral 8x7B-instruct\end{tabular} &
  88.14\% &
  34.81\% &
  55.55\% &
  11.11\% &
  29.62\% &
  66.66\% &
  8.88\% &
  21.48\% &
  77.77\% &
  11.11\% &
  41.85\% \\
\begin{tabular}[c]{@{}c@{}}GPT-3.5 Turbo\end{tabular} &
  77.77\% &
  47.41\% &
  86.66\% &
  {\color[HTML]{CB0000} \textbf{17.77\%}} &
  42.22\% &
  97.77\% &
  2.22\% &
  41.48\% &
  91.11\% &
  2.22\% &
  51.35\% \\
\begin{tabular}[c]{@{}c@{}}GPT-4 Turbo\end{tabular} &
  99.25\% &
  {\color[HTML]{CB0000} \textbf{93.33\%}} &
  82.22\% &
  {\color[HTML]{CB0000} \textbf{17.77\%}} &
  {\color[HTML]{CB0000} \textbf{92.59\%}} &
  {\color[HTML]{CB0000} \textbf{100\%}} &
  {\color[HTML]{CB0000} \textbf{15.55\%}} &
  {\color[HTML]{CB0000} \textbf{85.18\%}} &
  {\color[HTML]{CB0000} \textbf{100\%}} &
  {\color[HTML]{CB0000} \textbf{15.55\%}} &
  {\color[HTML]{CB0000} \textbf{80.12\%}} \\ \midrule
GPTZero &
  37.03\% &
  {\color[HTML]{303498} \textbf{100\%}} &
  6.66\% &
  0\% &
  {\color[HTML]{303498} \textbf{100\%}} &
  2.22\% &
  4.44\% &
  {\color[HTML]{303498} \textbf{100\%}} &
  0\% &
  0\% &
  56.91\% \\
Prepostseo &
  5l.85\% &
  97.03\% &
  20\% &
  2.22\% &
  96.29\% &
  17.77\% &
  2.22\% &
  95.55\% &
  4.44\% &
  6.66\% &
  60.2\% \\
WWW'23 detector &
  {\color[HTML]{303498} \textbf{82.22\%}} &
  80\% &
  {\color[HTML]{303498} \textbf{88.88\%}} &
  {\color[HTML]{303498} \textbf{82.22\%}} &
  77.77\% &
  {\color[HTML]{303498} \textbf{86.66\%}} &
  {\color[HTML]{303498} \textbf{84.44\%}} &
  83.7\% &
  {\color[HTML]{303498} \textbf{77.77\%}} &
  {\color[HTML]{303498} \textbf{88.88\%}} &
  {\color[HTML]{303498} \textbf{82.34\%}} \\ \bottomrule
\end{tabular}
}
\caption{Plagiarism type classification (original (i.e., `No') vs. verbatim (i.e., `Verb') vs. paraphrase (i.e., `Para') vs. summary (i.e., `Summ') plagiarism) performance of 8 detectors w.r.t. the models used for generation. For non-LLM approaches, we compute the category-wise breakdown from binary classification as they are not suitable for this task. `Acc' in the rightmost column indicates the overall accuracy regardless of category types. The highest values among LLM-based approaches are highlighted in {\color[HTML]{CB0000} \textbf{red}}, while the highest values for non-LLM-based approaches are highlighted in {\color[HTML]{303498} \textbf{blue}}.}
\label{tab:4way_perf}
\end{table*}

\subsection{RQ2. how well can LLMs detect plagiarism?}
\noindent \textbf{Binary Plagiarism Detection Results.} Figure \ref{fig:binary_perf} shows the plagiarism detection performance of $D_{src}$ and $D_{susp}$ in a binary setting (original vs. plagiarism). The results of zero-shot vanilla prompting, the most basic setup, hint that Llama2-70b-chat and GPT-3.5 Turbo perform almost close to a random guess (F1 = 0.5). According to our qualitative inspection, the errors from Llama2-70b-chat result from the model not being able to predict a final binary output. 26.7\% of Llama2-70b-instruct’s output does not provide the prediction at all. Additionally, its prediction tends to be highly skewed to the negative class. GPT-3.5 Turbo, on the other hand, generally can generate a binary prediction. However, it suffered from low recall (0.3). Comparing across the same families (Llama2 vs. Llama3 and GPT-3.5 vs. GPT-4), there is a significant performance gap. These findings are reasonable in the sense that Llama3 and GPT-4 are upgraded versions of Llama2 and GPT-3.5, respectively. A similar phenomenon has been reported in the task of medical final examination \cite{rosol2023evaluation}. We also find that some models achieve higher performance with CoT prompting and few-shot learning compared to zero-shot vanilla prompting. These results are consistent with previous literature \cite{wei2022chain, brown2020language}. In particular, Llama3-70b-instruct and GPT-4 Turbo achieve near-perfect performance. Yet, the remaining models often obtain incorrect responses due to hallucination. For instance, although their reasoning steps are logically valid, their final response contradicts them.

Commercial plagiarism detectors, GPTZero and Prepostseo, are shown to achieve very low F1 scores. This may be due to two reasons: (1) their built-in retrieval systems fail to fetch relevant documents from the Internet; these tools rely on simple string-matching algorithms designed specifically for detecting verbatim plagiarism, rather than accommodating a broader range of plagiarism categories. Conversely, the WWW'23 detector exhibits stronger performance as it is specifically tailored for this particular task, extrinsic plagiarism detection. Nonetheless, four LLMs have been shown to surpass their performance through few-shot CoT prompting. Overall, these results highlight the potential of LLMs in effectively detecting plagiarism.

\vspace{5pt}
\noindent \textbf{Plagiarism Type Classification Results.} 
Table \ref{tab:4way_perf} shows detection results of 8 detectors on plagiarism type classification depending on plagiarism types and models used for generation. Of the four prompting techniques, we resort to zero-shot CoT prompting for this experiment. This decision is motivated by the fact that its inference time is significantly faster than few-shot CoT prompting with improved detection performance. Here, due to category-wise performance computation, we report accuracy scores. In line with the results from binary plagiarism detection, Llama3-70b-instruct, and GPT-4 Turbo are the highest-performing LLMs. In particular, their identification of verbatim and paraphrase plagiarism achieved 99-100\% accuracy. GPT-3.5 Turbo is demonstrated to be quite capable of distinguishing paraphrase plagiarism, but it performs poorly in identifying plagiarism-free content and summarized content. Most notably, LLMs perform poorly in detecting summary plagiarism across all genres. Most of their errors stem from short texts being confused for paraphrase plagiarism instead of summary plagiarism. 

Now, looking at traditional plagiarism checkers, we find that GPTZero and Prepostseo excel at identifying non-plagiarized content compared to the WWW'23 detector. Yet, the WWW'23 detector surpasses them in paraphrase and summary detection. This discrepancy may be due to differing definitions of plagiarism; since these tools are designed for detecting verbatim plagiarism, they may incorrectly classify text pairs with minimal lexical overlap as non-plagiarized.

\section{Conclusion}
We present a novel large-scale plagiarism detection benchmark, {\sf PlagBench}: a collection of
46.5K artificial plagiarism cases generated by three cutting-edge LLMs. This underscores the nuanced challenges and opportunities presented by LLMs in combating plagiarism while also acknowledging their potential vulnerabilities to being used for unethical practices. We envision {\sf PlagBench} as a universal standard for evaluating plagiarism detection methods.


\section*{Limitations}
The current research has two limitations. First, we use a single manually crafted prompt template for generating machine-plagiarized documents and detecting plagiarism. There are, however, various prompting techniques and strategies for automatically optimizing prompts. We expect future work to explore a broader range of prompts for both generation and detection purposes. Second, our experiments are based on instruction-tuned decoder-only LLMs. It is uncertain whether the observed findings would apply to other types of LLMs with different architectures. Future research should revisit our RQs using more diverse model types.

\section*{Ethics Statement}
This research involves the use of LLMs to simulate plagiaristic behavior. All source documents utilized in this study are derived from publicly accessible open-source datasets. To promote transparency and reproducibility, we will make all data and code used in our experiments available to the research community. We acknowledge the potential ethical concerns associated with generating synthetic plagiarism documents. Therefore, we emphasize that these documents should be used exclusively for research purposes aimed at understanding and mitigating plagiarism. We advise against any misuse of the synthetic data, such as using it to deceive or harm others. Our work is intended to contribute to the development of more effective plagiarism detection tools and to enhance academic integrity. By sharing our resources, we aim to support the broader research community in these endeavors.

Regarding human annotation, our research protocol was approved by the Institutional Review Board (IRB) at our institution. We only recruited annotators that are 18 years old or over. All annotators were paid over minimum wage rate.

\section*{Acknowledgments}
This work was in part supported by NSF awards \#1934782 and \#2114824. Some of the research results were obtained using computational resources provided by CloudBank (\url{https://www.cloudbank.org/}), which was supported by NSF award \#1925001. In addition, this work is partially supported by DHS (17STQAC00001-07-00). The views and conclusions contained in this paper are those of the authors and should not be interpreted as representing any funding agencies.

\bibliography{acl}

\appendix

\section{Appendix}
\subsection{Automatic Evaluation of LLM-Generated Plagiarism}
\label{sec:appendix_threshold}
We leverage publicly available two datasets, PAWS \cite{paws2019naacl} and SummEval \cite{fabbri2021summeval}, to find the optimal thresholds for data filtering regarding paraphrase evaluation and summary evaluation, respectively. Both datasets contain human annotation labels signaling whether a pair of texts is correct paraphrases or not. PAWS provide one binary label for each text pair, whereas SummEval contains a list of 8 annotation results in which the scores range from 1 to 5 evaluated across 4 dimensions: coherence, consistency, fluency, and relevance. 

The threshold selection process is straightforward; we apply our established automated evaluation measurements to the subset of these two corpora and compute their mean scores of \textit{bad} examples. Specifically, we compute mean scores of entries with `0' label from the PAWS dataset and entries with average annotation scores lower than 2.5 from the SummEval dataset. These mean scores are then used as a reference to assign filtering thresholds. 
Give two pieces of evaluation texts $(d, d^*)$, let's denote paraphrase evaluation results as $R_{\text{para\_eval}}$. $R_{\text{para\_eval}}$ can be expressed as below:  

{\small
\begin{align*}
R_{\text{para\_eval}} = 
\begin{cases}
    1 & \text{if } \text{Alignscore}(d, d^*) > 0.5 \\
      & \quad \land \text{ BERTScore}(d, d^*) > 0.5 \\
      & \quad \land \text{ Readability\_Gap}(d, d^*) > -0.5 \\
      & \quad \land \text{ COLAscore\_Improv}(d, d^*) \geq 0 \\
    0 & \text{else}.
\end{cases}
\end{align*}
}

\noindent Summary evaluation results are denoted as $R_{\text{summ\_eval}}$.
{\small
\begin{align*}
R_{\text{summ\_eval}} = 
\begin{cases}
    1 & \text{if } \text{Alignscore}(d, d^*) > 0.5 \\
      & \quad \land \text{ BLANC}(d, d^*) > 0.0 \\
      & \quad \land \text{ BARTScore}(d, d^*) > -4.0 \\
      & \quad \land \text{ Readability}(d, d^*) > 10 \\
      & \quad \land \text{ COLAscore\_Improv}(d, d^*) \geq 0 \\
    0 & \text{else}.
\end{cases}
\end{align*}
}

Given these two definitions, we omit all samples associated with `0' in regards to $R_{\text{para\_eval}}$ and $R_{\text{summ\_eval}}$.

\subsection{Details on Human Evaluation of LLM-Generated Plagiarism}
\label{sec:appendix_amt}
\textbf{Task Formation.} Each task consists of three batches of one source text and three suspicious texts generated by Llama2-70b-chat, GPT-3.5 Turbo, and GPT-4 Turbo. Given a pair of the source text and machine-paraphrased or machine-summarized texts, we first verbally describe our established evaluation aspects on the task description and ask them to answer if the described aspect is satisfied or not. For plagiarism-free cases, the definitions of paraphrase and summary plagiarism are provided, and annotators are tasked to identify two texts belong to either paraphrase or summary plagiarism. See Table \ref{tab:evaluation_question1}, and \ref{tab:evaluation_question2}, and \ref{tab:evaluation3}. Given that the estimated completion time per batch is 10 minutes, workers are compensated $\$1.5$ per batch based on United States average hourly wages.

\subsection{Model Size And Budget}
We enlist model size used for our experiment here: for Llama2 and Llama3, we use the largest scale among their models, which is 70B. GPT-3.5 is expected to contain 175B parameters, equivalent to GPT-3. Meanwhile, GPT-4 has 1.76T parameters, and Mixtral has 46.7B total parameters. For all experiments, we leverage API calls from Huggingface and OpenAI official website.

\newpage

\begin{table*}[htbp]
\footnotesize{
\centering
\begin{tabular}{p{16cm}} 
\toprule
\textbf{Instruction} \\ \midrule
In this HIT, you will be presented with 5 paragraphs consisting of 1 source text and 4 machine-generated texts covering a similar topic as the source text. In total, there will be 3 sets of 5 paragraphs for you to evaluate. Please read the texts carefully and answer the following questions to evaluate if the provided texts plagiarize the source text. There are 2 plagiarism types you need to consider: \\ 

• Paraphrase plagiarism: The evaluation text is rephrased or rewritten using different words but retains the same meaning and structure as the source text. \\
• Summary plagiarism: The evaluation text encapsulates the most essential points of the source text into a shorter form. \\ \midrule
\textbf{FIRST SET} \\
Source Text: \texttt{\$\{source\_doc\}} \\ 
Evaluation Text A: \texttt{\$\{evaluation\_doc\}} \\ \midrule

\textbf{Questions for Evaluation Text A} \\
\textbf{Q1. Paraphrase plagiarism}: Is the evaluation text a paraphrase of the source text (i.e., rephrased or rewritten using different words but retaining the same meaning and structure)? \\
\textbf{Answer}: Yes/No \\ 
\textbf{Q2. Summary plagiarism}: Does the evaluation text summarize the source text (i.e., encapsulate the most essential points of the source text into a shorter form)? \\
\textbf{Answer}: Yes/No \\ 
\bottomrule
\end{tabular}
\caption{Human annotation instruction for plagiarism-free cases}
\label{tab:evaluation_question1}
}
\end{table*}


\begin{table*}[htbp]
\footnotesize{
\centering
\begin{tabular}{p{16cm}} 
\toprule
\textbf{Instruction} \\ \midrule
In this HIT, you will be presented with 5 paragraphs consisting of 1 source text and 4 corresponding machine-generated texts. In total, there will be 3 sets of 5 paragraphs for you to evaluate. Please read the texts carefully and answer the following questions to evaluate the overall quality of the provided texts given the source text. \\ \midrule

\textbf{FIRST SET} \\ 
Source Text: \texttt{\$\{source\_doc\}} \\ 
Evaluation Text A: \texttt{\$\{evaluation\_doc\}} \\ \midrule

\textbf{Questions for Evaluation Text A} \\ 
\textbf{Q1. Relevance}: Does the evaluation text cover all the essential information and key points (without including irrelevant or tangential details) from the source text? \\
\textbf{Answer}: Yes/No \\ 
\textbf{Q2. Consistency}: Does the evaluation text accurately represent the facts, details, and information contained in the source text? \\
\textbf{Answer}: Yes/No \\ 
\textbf{Q3. Coherence}: Is the evaluation text well-structured and not just a random assortment of information from the source text? \\
\textbf{Answer}: Yes/No \\ 
\bottomrule
\end{tabular}
\caption{Human annotation instruction for paraphrase plagiarism cases}
\label{tab:evaluation_question2}
}
\end{table*}




\begin{table*}[htbp]
\footnotesize{
\centering
\begin{tabular}{@{}p{15cm}@{}}
\toprule
\textbf{Instruction} \\ \midrule
In this HIT, you will be presented with 5 paragraphs consisting of 1 source text and 4 corresponding machine-generated texts. In total, there will be 3 sets of 5 paragraphs for you to evaluate. Please read the texts carefully and answer the following questions to evaluate the overall quality of the provided texts given the source text. \\ \midrule

\textbf{FIRST SET} \\ 
Source Text: \texttt{\$\{source\_doc\}} \\ 
Evaluation Text A: \texttt{\$\{evaluation\_doc\}} \\ 

\textbf{Questions for Evaluation Text A} \\ 
\textbf{Q1. Relevance}: Does the evaluation text cover all the essential information and key points (without including irrelevant or tangential details) from the source text? \\
\textbf{Answer}: Yes/No \\ 
\textbf{Q2. Consistency}: Does the evaluation text accurately represent the facts, details, and information contained in the source text? \\
\textbf{Answer}: Yes/No \\ 
\textbf{Q3. Coherence}: Is the evaluation text well-structured and not just a random assortment of information from the source text? \\
\textbf{Answer}: Yes/No \\ 
\textbf{Q4. Language quality}: Does the evaluation text maintain or improve upon the fluency and grammaticality of the source text? \\
\textbf{Answer}: Yes/No \\ \bottomrule
\end{tabular}
\caption{Human annotation instruction for summary plagiarism cases}
\label{tab:evaluation3}
}
\end{table*}

\begin{table*}[]
\footnotesize{
\setlength{\tabcolsep}{0.5em} 
{\renewcommand{\arraystretch}{1.3}
\centering
\begin{tabular}{|c|c|c|c|c|}
\hline
\textbf{Domain} &
  \textbf{Model} &
  \textbf{\# of document pairs} &
  \textbf{\begin{tabular}[c]{@{}c@{}}\# of remaining pairs \\ (paraphrase)\end{tabular}} &
  \textbf{\begin{tabular}[c]{@{}c@{}}\# of remaining pairs \\ (summary)\end{tabular}} \\ \hline
\multirow{3}{*}{\textbf{SciXGen}}    & Llama2-70b-chat & 2,400 & 1,179 (49.12\%) & 2,147 (89.46\%) \\
                                     & GPT-3.5 Turbo   & 2,400 & 1,764 (73.5\%)  & 2,379 (99.12\%) \\
                                     & GPT-4 Turbo     & 1,300 & 1,085 (83.46\%) & 1,266 (97.38\%) \\ \hline
\multirow{3}{*}{\textbf{ROCStories}} & Llama2-70b-chat & 2,400 & 1,699 (70.79\%) & 1,975 (82.29\%) \\
                                     & GPT-3.5 Turbo   & 2,400 & 2,385 (99.38\%) & 1,614 (67.25\%) \\
                                     & GPT-4 Turbo     & 1,300 & 1,295 (99.62\%) & 1,166 (89.69\%) \\ \hline
\multirow{3}{*}{\textbf{TLDR}}       & Llama2-70b-chat & 1300  & 753 (57.92\%)   & 1,060 (81.54\%) \\
                                     & GPT-3.5 Turbo   & 1300  & 1,230 (99.38\%) & 1,181 (90.85\%) \\
                                     & GPT-4 Turbo     & 700   & 681 (97.29\%)   & 657 (93.86\%)   \\ \hline
\end{tabular}
}
}
\caption{Dataset statistics after automatic filtering for paraphrase and summary plagiarism. The percentage in the bracket represents the percentage of remaining document pairs remaining out of original pairs.}
\label{tab:filter_stats}
\end{table*}

\begin{table*}[]
\footnotesize{
\setlength{\tabcolsep}{0.5em} 
{\renewcommand{\arraystretch}{1.3}
\centering
\begin{tabular}{|c|c|c|c|c|}
\hline
\textbf{Domain}                       & \textbf{Model}                         & \textbf{\begin{tabular}[c]{@{}c@{}}\# of document \\ (plagiarism-free)\end{tabular}} & \textbf{\begin{tabular}[c]{@{}c@{}}\# of document\\ (paraphrase)\end{tabular}} & \textbf{\begin{tabular}[c]{@{}c@{}}\# of document \\ (summary)\end{tabular}} \\ \hline
                                      & {\color[HTML]{333333} Llama2-70b-chat} & 108                                                                                  & 83                                                                             & 102                                                                          \\
                                      & GPT-3.5 Turbo                          & 99                                                                                   & 87                                                                             & 132                                                                          \\
\multirow{-3}{*}{\textbf{SciXGen}}    & GPT-4 Turbo                            & 118                                                                                  & 45                                                                             & 129                                                                          \\ \hline
                                      & {\color[HTML]{333333} Llama2-70b-chat} & 196                                                                                  & 55                                                                             & 118                                                                          \\
                                      & GPT-3.5 Turbo                          & 196                                                                                  & 45                                                                             & 120                                                                          \\
\multirow{-3}{*}{\textbf{ROCStories}} & GPT-4 Turbo                            & 197                                                                                  & 56                                                                             & 117                                                                          \\ \hline
                                      & {\color[HTML]{333333} Llama2-70b-chat} & 114                                                                                  & 55                                                                             & 97                                                                           \\
                                      & GPT-3.5 Turbo                          & 107                                                                                  & 56                                                                             & 114                                                                          \\
\multirow{-3}{*}{\textbf{TLDR}}       & GPT-4 Turbo                            & 113                                                                                  & 45                                                                             & 129                                                                          \\ \hline
\end{tabular}
}
}
\caption{Dataset statistics after manual filtering for plagiarism labels.}
\label{tab:human_filter_stats}
\end{table*}

\begin{table*}[]
\definecolor{apricot}{RGB}{242, 157, 96}
\footnotesize{
\setlength{\tabcolsep}{0.5em} 
{\renewcommand{\arraystretch}{1.3}
\begin{tabular}{|c|l|}
\hline
\textbf{Plagiarism category} & \multicolumn{1}{c|}{\textbf{Template}}                                                                                                                                                                                                                                                                          \\ \hline
\textbf{Paraphrase}          & {\color[HTML]{333333} \begin{tabular}[c]{@{}l@{}}Paraphrase the following text while keeping its meaning.\\ \\ Text: \texttt{\$\{source\_doc\}} \\ Paraphrased:\end{tabular}}                                                                                                                                           \\ \hline
\textbf{Summary}             & \begin{tabular}[c]{@{}l@{}}Summarize the following text in 1-3 sentences.\\ \\ Text: \texttt{\$\{source\_doc\}}  \\ Summarized:\end{tabular}                                                                                                                                                                             \\ \hline
\textbf{Plagiarism-Free}     & \begin{tabular}[c]{@{}l@{}}Based on the provided text passage and keywords, write its continuation in a style of \texttt{\$\{genre\}} .\\ When generating, make sure that the continuation is coherent while including all keywords.\\ \\ Text: \texttt{\$\{start\_sent\}} \\ Keywords: \texttt{\$\{keyword\}} \\ Continuation:\end{tabular} \\ \hline
\end{tabular}
}
}
\caption{Prompt templates for plagiarism generation task.}
\label{tab:prompt_template_gen}
\end{table*}

\begin{table*}[]
\definecolor{apricot}{RGB}{242, 157, 96}
\resizebox{\textwidth}{!}{%
\begin{tabular}{|c|l|}
\hline
\textbf{Prompt type} &
  \multicolumn{1}{c|}{\textbf{Template}} \\ \hline
\textbf{\begin{tabular}[c]{@{}c@{}}Zero-shot vanilla\\ prompting \\ (binary detection)\end{tabular}} &
  {\color[HTML]{000000} \begin{tabular}[c]{@{}l@{}}There are three types of plagiarism:\\ • Verbatim plagiarism: the evaluation text consists of exact copies of words or phrases \\ without transformation from the source text without citation.\\ • Paraphrase plagiarism: the evaluation text is rephrased or rewritten using different words but \\ retain the same meaning and structure as the source text without citation.\\ • Summary plagiarism: the evaluation text encapsulates the most essential points of the source text \\ into a shorter form without citation.\\ \\ If the evaluation text does not belong to any of three plagiarism categories, it means plagiarism-free.\\ Given a pair of text and provided plagiarism definitions, does the evaluation text plagiarize the source text \\ (yes/no)? Please format your final response as ‘Answer: \{\{response\}\}’.  \\ \\ Source text: \texttt{\$\{source\_doc\}}\\ Evaluation text: \texttt{\$\{evaluation\_doc\}}\end{tabular}} \\ \hline
\textbf{\begin{tabular}[c]{@{}c@{}}Zero-shot CoT\\ prompting\\ (binary detection)\end{tabular}} &
  \begin{tabular}[c]{@{}l@{}}There are three types of plagiarism:\\ • Verbatim plagiarism: the evaluation text consists of exact copies of words or phrases \\ without transformation from the source text without citation.\\ • Paraphrase plagiarism: the evaluation text is rephrased or rewritten using different words but \\ retain the same meaning and structure as the source text without citation.\\ • Summary plagiarism: the evaluation text encapsulates the most essential points of the source text\\  into a shorter form without citation.\\ \\ If the evaluation text does not belong to any of three plagiarism categories, it means plagiarism-free.\\ Given a pair of text and provided plagiarism definitions, does the evaluation text plagiarize the source text\\ (yes/no)?  First think step-by-step and format your final response as ‘Final answer: \{\{response\}\}’.\\ \\ Source text: \texttt{\$\{source\_doc\}}\\ Evaluation text: \texttt{\$\{evaluation\_doc\}}\\ \\ Answer: Let's think step-by-step.\end{tabular} \\ \hline
\textbf{\begin{tabular}[c]{@{}c@{}}Zero-shot vanilla\\ prompting\\ (type identification)\end{tabular}} &
  \begin{tabular}[c]{@{}l@{}}There are three types of plagiarism:\\ • Verbatim plagiarism: the evaluation text consists of exact copies of words or phrases\\  without transformation from the source text without citation.\\ • Paraphrase plagiarism: the evaluation text is rephrased or rewritten using different words but \\ retain the same meaning and structure as the source text without citation.\\ • Summary plagiarism: the evaluation text encapsulates the most essential points of the source text \\ into a shorter form without citation.\\ • No plagiarism: the evaluation text does not belong to any of three plagiarism categories.\\ \\ Given a pair of text and provided plagiarism definitions, what type of plagiarism \\ (no/verbatim/paraphrase/summary) does the evaluation belong to when compared to the source text? \\ Please format your final response as ‘Answer: \{\{response\}\}’.  \\ \\ Source text: \texttt{\$\{source\_doc\}}\\ Evaluation text: \texttt{\$\{evaluation\_doc\}}\end{tabular} \\ \hline
\textbf{\begin{tabular}[c]{@{}c@{}}Zero-shot CoT\\ prompting\\ (type identification)\end{tabular}} &
  \begin{tabular}[c]{@{}l@{}}There are three types of plagiarism:\\ • Verbatim plagiarism: the evaluation text consists of exact copies of words or phrases\\  without transformation from the source text without citation.\\ • Paraphrase plagiarism: the evaluation text is rephrased or rewritten using different words but \\ retain the same meaning and structure as the source text without citation.\\ • Summary plagiarism: the evaluation text encapsulates the most essential points of the source text \\ into a shorter form without citation.\\ • No plagiarism: the evaluation text does not belong to any of three plagiarism categories.\\ \\ Given a pair of text and provided plagiarism definitions, what type of plagiarism\\ (no/verbatim/paraphrase/summary)  does the evaluation belong to when compared to the source text? \\ First think step-by-step and\\  format your final response as ‘Answer: \{\{response\}\}’.  \\ \\ Source text: \texttt{\$\{source\_doc\}}\\ Evaluation text: \texttt{\$\{evaluation\_doc\}}\\ \\ Answer: Let's think step-by-step.\end{tabular} \\ \hline
\end{tabular}
}
\caption{Prompt templates for plagiarism detection task.}
\label{tab:prompt_example}
\end{table*}

\end{document}